# TAXONOMY, STRUCTURE AND IMPLEMENTATION OF EVIDENTIAL REASONING MODELS

Moshe Ben-Bassat

Faculty of Management  and  University of California
Tel Aviv University         Computer Science Department
Tel Aviv 69978              Los Angeles, California 90024
ISRAEL                      U.S.A.
B25@TAUNIVM.BITNET

ABSTRACT

The fundamental elements of evidential reasoning problems are described, followed by a discussion of the structure of various types of problems. Bayesian inference networks and state space formalism are used as the tool for problem representation.

A human-oriented decision making cycle for solving evidential reasoning problems is described and illustrated for a military situation assessment problem. The implementation of this cycle may serve as the basis for an expert system shell for evidential reasoning; i.e. a situation assessment processor.

Keywords: Evidential reasoning, Bayesian inference networks. Expert ststems, situation assessment.

Acknowledgement: This work was supported in part by the National Science Foundation Grant #DSR 83-13875.
   The author is grateful to Dr. Judea Pearl for many useful discussions.

## 1. PROBLEM STATEMENT

Evidential reasoning refers to inference mechanisms by which the evidence provided by a set of indicators (findings, features, attributes, variables) is analyzed in order to gain better understanding of a given hypothesis, concept, situation, or phenomenon. Problem solving tasks of this nature include, for instance, medical diagnosis,, weather forecasting, corporate assessment, political crisis assessment, and battlefield reading. All of these problems share common characteristics in that the problem solver (PS) starts with an initial incomplete understanding of the situation (e.g. patient status) and based on his prior knowledge and expectations, he then looks for additional information that may reduce the uncertainty regarding the complete picture of the situation. Following a cyclic process, sources for additional information are identified, evaluated, and utilized, the new evidence is integrated into the existing knowledge base, the situation is reassessed and, if final assessments cannot be made, further information is requested. The process ends when the problem solver (PS) decides that he knows enough to form a defensible interpretation of the situation, or that no additional
sources of information can contribute significantly (compared to
their cost) to remove the uncertainty which still remains, or
temporal considerations force him to terminate information
acquisition and assess the situation as best he can.

Models for evidential reasoning and uncertainty management
have attracted significant scientific effort since the beginning
of the century. Classical probability theory, Carnap's and
Hempel's confirmation theory, Shafer's evidence theory, and
Zadeh's possibility theory represent a sample of such works. Many
of the classical models were adopted and improved by artificial
intelligence researchers who applied them in a variety of expert
systems such as MYCIN (Shortliffe 1976), PROSPECTOR (Duda 1979)
and MEDAS (Ben-Bassat 1980).

The purpose of this paper is to present a draft taxonomy of
evidential reasoning problems and to propose a framework by which
evidential reasoning models may be evaluated and compared. As a
frame of reference we propose models which are based on Bayesian
(probabilistic) inference networks.

## 2. PROBLEM AND KNOWLEDGE REPRESENTATION

### Bayesian inference Networks

Problem and knowledge representation for evidential reasoning
tasks may be based on uncertain hierarchical inference networks.
In such networks leaf nodes typically represent observable events
(indicators), while higher level nodes represent events
(hypotheses) whose value (true, false or other) may be inferred
from other nodes in the network; typically in lower levels but not
necessarily.

Formally, a node represents a multi-valued proposition in
which the values are mutually exclusive and exhaustive. If this is
not the case, we break the node into separate nodes each of which
represents a set of mutually exclusive propositions.

At any given point of time, we assign to each node a set of
values that correspond to our degree of belief in the validity of
the alternative propositions represented by that node. In Bayesian
networks node values stand for the probabilities of the various
alternatives corresponding to that node.

A link between nodes $H_i$ and $E_j$ represents evidential
relevancy between the two corresponding events. Each link is
assigned value(s) that represent the degree of significance for
inferring $H_i$ from $E_j$ or vice versa. In Bayesian networks a
directed link that emanates from $H_i$ pointing at $E_j$ is assigned a
matrix that represents $P(e_j \mid h_i)$ for all of the possible values
of $H_i$ and $E_j$. By this formulation we are not commiting ourselves
whether the link represents a causal relationship
(i.e. P(sympton|disease)) or a diagnostic relationship
(i.e. P(disease|symptom)). It is our experience, however, that in
most cases eliciting causal probabilities is preferable. See

18

* (Ben-Bassat 1980) p.150. for a discussion in the context of medical diagnosis.

Once an observable node is reported its evidence is propagated along the network links and revises our belief in the validity of the higher level hypotheses connected to that node. In Bayesian networks, propagation mechanisms are based on Bayes theorem as the fundamental tool for probability revision, e.g. (Pearl 1986a).

### Node Categorization

The hierarchical network structure suggests a categorization of the nodes into three main types. The leaf nodes typically represent events that can be perceived directly by the system sensors (the "eyes", "ears" and the keyboard). Higher level nodes typically represent events that are deduced by the system inference engine ("brain"). Root nodes represent the target hypotheses whose resolution is the ultimate objective of the system. Intermediate nodes may or may not be on the list of target hypotheses (see below), and, in any case, we use them to form defensible argumentations of the resolution of higher level hypotheses.

Several comments, however, are in order:
1. An intermediate or top level node may sometimes be directly observable, but at a higher cost than inferring it from observable lower level indicators. For instance, opening the abdomen (explorative laparotomy) provides direct observation on events that we initially attempt to deduce from less expensive observations.

2. We may sometimes wish to bypass low level nodes and report a value directly into an intermediate or top level node. This value is not an observation but rather a deduction performed by an autonomous agent who is unable or prefers not to delineate the basis for his deduction by lower level nodes. An example would be a distributed military intelligence operation in which medium level officers report upward only their summarized assessments.

3. An observable indicator may sometimes be observed with noise. In this case we report upward a set of probabilities that summarize our impression of the noisy observation with regard to the possible values of the node. An example would be a patient who does not respond unequivocally to a physician's questions.

4. Although the structure indicates that evidence is propageted bottom up; top-down and sideways propagation may sometimes be found very useful. In fact, an important feature of Bayesian propagation is that it permits propagation in all directions.

### State Space Representation

Using this framework, we may represent evidential reasoning tasks by a state space formalism. The state of the system at any given stage is characterized by the current values on the network nodes. For the initial state $S_0$ we assign to all top level
hypotheses their prior probabilities. The initial values for
intermediate and leaf nodes may be derived from their parents and
the values on the links. Observable nodes that have not yet been
observed are additionally assigned the value UNOBSERVED designated
by "?". These nodes are candidates for direct observation to be
suggested by the information acquisition process.

From the goal state point of view, the network nodes are
divided into target and non-target nodes. Target nodes represent
hypotheses that need to be resolved by the end of the process.
That is upon process termination we need to make a commitment on a
set of values - one for each target node - that jointly constitute
the best explanation for the existing evidence.

The set of target nodes depends on the application. In some
cases the only important decisions at the final stage are about
the root nodes and decisions about other nodes do not lead to any
operational consequences. In these cases the goal state depends on
the values of the root nodes only.

In other cases the state of intermediate nodes impact the
action plan (in addition to their role as mediators for higher
level deductions). For instance, in medical diagnosis of critical
care disorders a node representing the state of SHOCK is an
intermediate node. Yet, to device a treatment plan it is very
important to know whether the patient is or is not in SHOCK.

<u>Optimal</u> termination criteria and commitment rules are complex
issues which are still in their infancies for probabilistic
inference networks. See however; (Ben-Bassat 1980b) and (Pearl
1986b). An example of a simplified goal state for medical
diagnosis is as follows:

$S_G$: The values of the top level hypotheses(is) and a selected
group of intermediate hypotheses are above or below certain
thresholds.

A more sophisticated state space formulation of diagnostic
problems is presented by (Ben-Bassat 1985a).

Evidential reasoning is the process of transferring the
network from its initial state $S_0$ to a goal state $S_G$. The
operators for this transformation are queries on the observable
nodes. The objective of a control strategy (Ben-Bassat 1985b) is
to reach a goal state in a cost-effective manner.

3. <u>TYPES OF EVIDENTIAL REASONING PROBLEMS</u>

Three main factors play a role in determining the difficulty
of evidencial reasoning problems:

1. Network structure (depth, width, loops...)

2. Target nodes; their number and interrelationships among them.

20

3. Dependencies among observable nodes.

In what follows we will describe several types of evidential reasoning problems based on the first two factors only. These are graphically illustrated in Figure 1 through Figure 6 where the following notation is used:

observable node
target node
intermediate node

Case (a) (Figure 1)
- One set of hypotheses which are mutually exclusive and exhaustive.
- Observations which are directly linked to the hypotheses

Case (a) is a representative of the well known classical Bayesian classification problem that has been extensively researched in statistics, decision theory and pattern recognition.

Case (b) (Figure 2)

- One set of hypotheses which are mutually exclusive and exhaustive.
- Hierarchical tree-like inferential links.

Reasearchers in behavioral decision theory refer to case (b) as cascaded inference; (Schum 1978). (The example in Figure 2 was given by J. Pearl).

Case (c) (Figure 3)

- Hierarchically structured mutually exclusive and exhaustive hypotheses
- Observations which are directly linked to groups of hypotheses at differet layers of the hierarchy.

An example of case (c) appears in threat assessment of unknown objects. Some observable nodes may point directly into the root node that represents the general hypothesis whether the object is at all threatening or not. This hypothesis, however, represents several families of hypotheses concerning the various types of threats, each of which may, in turn, be subdivided into more refined classification up to point where each specific type of threat occupies a separate node. For each of the subfamilies we may have direct links from observable nodes and perhaps intermediate node. Obviously, if E provides evidence for a family H, then it also provides some evidence for all of the subfamilies of H. Gordon and Shortliffe (1985) and Pearl (1986) deal with this problem.

Case (d) (Figure 4)

- Multiple non-competing sets of hypotheses
- Observations which are directly linked to the hypotheses.

21

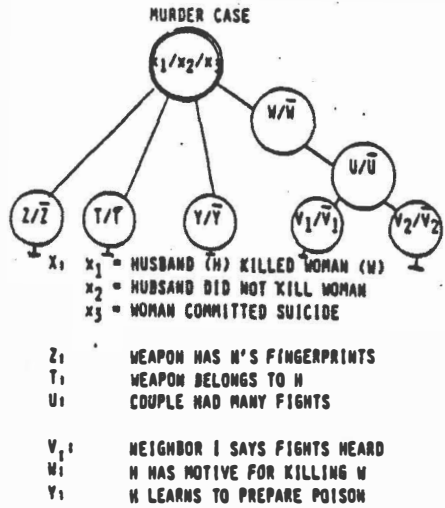

- $x_1$ = HUSBAND (H) KILLED WOMAN (W)
- $x_2$ = HUSBAND DID NOT KILL WOMAN
- $x_3$ = WOMAN COMMITTED SUICIDE

- $Z_1$   WEAPON HAS H'S FINGERPRINTS
- $T_1$   WEAPON BELONGS TO H
- $U_1$   COUPLE HAD MANY FIGHTS
- $V_1$:  NEIGHBOR 1 SAYS FIGHTS HEARD
- $W_1$   H HAS MOTIVE FOR KILLING W
- $Y_1$   H LEARNS TO PREPARE POISON

Figure 2

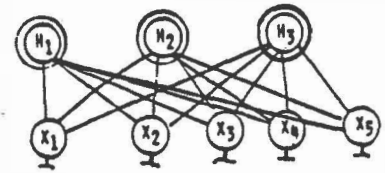

- A DEFENSIVE FORCE (D) CONTROLS THREE MAJOR POSTS $H_1$, $H_2$, $H_3$ OF A BATTLEFIELD
- AN OFFENSIVE FORCE (O) MAY ATTACK NONE, ONE, TWO OR ALL THREE OF THE POSTS
- D MAY TAKE RECONNAISSANCE ACTIONS IN ORDER TO REDUCE THE UNCERTAINTY. TYPICAL INDICATORS RESULTING FROM SUCH ACTIONS MAY INCLUDE:

$x_1$ - INCREASED ACTIVITY IN THE NORTHERN AREA
$x_2$ - BRIDGING EQUIPMENT MOVED FORWARD

Figure 3

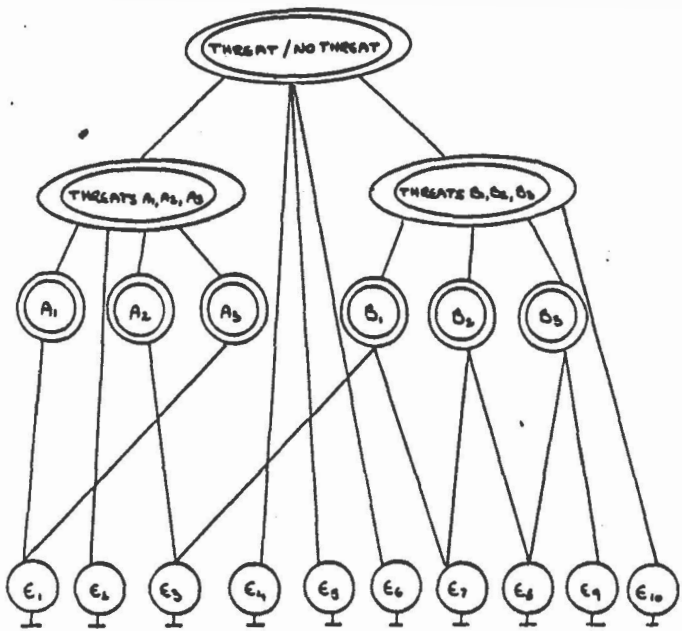

Figure 4

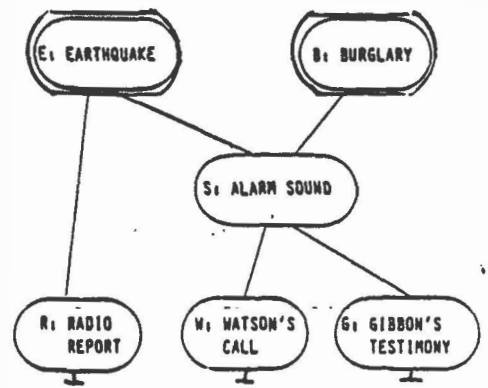

Figure 5

22

Case d refers to a situation where several hypotheses may co-exist simultaneously. In such a case every hypothesis Hi gets its own node with two possible values Hi and H̄i. The problem is also known as multi-membership classification in the context of pattern recognition (Ben-Bassat 1980b), and was recently addressed by (Pearl 1986c).

Case (e) (Figure 5)

- Multiple partially competing sets of hypothesis
- Multi-link inferential chains without loops (singly connected graph).

Case (e) represents a situation of multiple causes for a given observation. The example presented in Figure 5 was discussed by (Kim and Pearl 1983). The top level hypotheses are partially competing in the sense that "earthquake" reduces the liklihood of "burglary" by "explainig away" the alarm sound.

Case (f) (Figure 6)

- Multi-perspective hierarchical reasoning
  (Target nodes distributed all over the network)

In many applications, e.g. scene analysis, military situation assessment, we need to view an object or a situation from multiple perspectives in order to generate a rich description of it. For instance, in order to analyze a potential military attack (Ben-Bassat and Freedy 1982) we need to consider several perspectives: TYPE, THRUST, TARGET, TACTICS, DEPLOYMENT, etc. In each of these interrelated perspectives, the situation may be classified within one or more of the alternatives (states or classes) associated with that perspective. For example, an enemy attack can be one of the following TYPES: DELIBERATE, HASTY, SPOILING, or an AMBUSH. Similarly, there are several alternatives for THRUST, TARGET, TACTICS, etc.

Within a given perspective several alternatives may co-exist simultaneously. For instance, within the THRUST perspective there is no reason to assume a priori that the enemy attack will consist of TANKS only or PARACHUTES only. Any combination of the possible alternatives, TANKS, AIR, MOBILE INFANTRY, PARACHUTES, HELICOPTER CARRIED INFANTRY, may, in principle, be simultaneously true.

The recognition process is to some extent "hierarchical" in the sense that low-level indications are used as the building blocks of higher level indications. For instance, information regarding the presence of trees, their height and density, are features that contribute to determine COVER and CONCEALMENT. Boulder size and soil type contribute to determine tank TRAFFICABILITY. Together they contribute to TERRAIN analysis. The results of TERRAIN analysis and other factors such as CAPABILITY contribute, in turn, to the determination of what TACTICS the enemy may choose, his DEPLOYMENT technique, and even influence the choice of a TARGET.

23

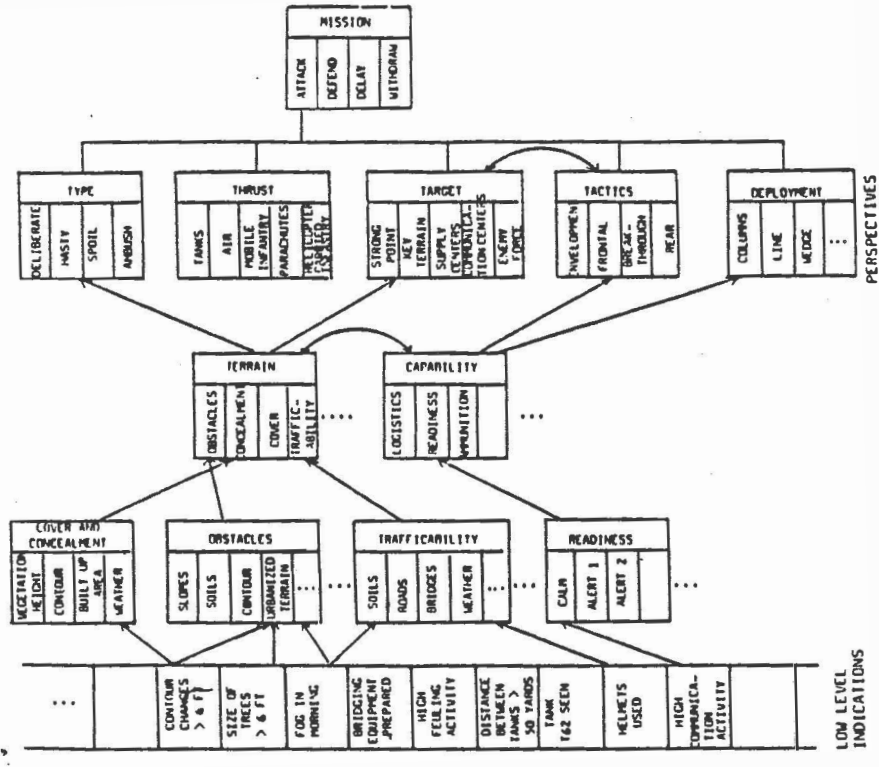

FIGURE 6. ANALYSIS OF MILITARY SITUATION ASSESSMENT

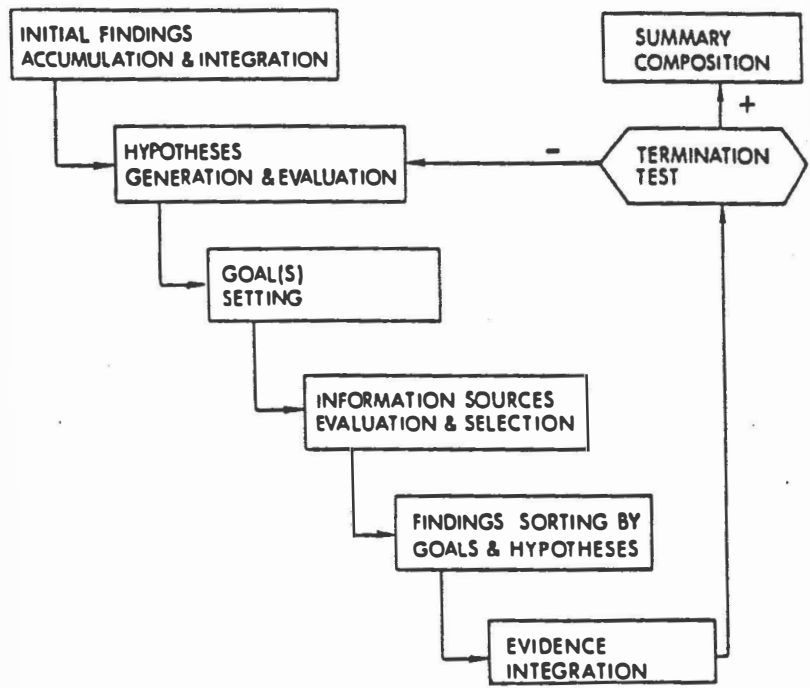

Figure 7. The cycle of evidential reasoning

24

The target nodes in this case are distributed all over the network because to device a battle plan we need to know the <u>details</u> of the arena and the enemy intentions, and not just whether or not he intends to attack.

## 4. IMPLEMENTATION IN EXPERT SYSTEMS

Expert systems for evidential reasoning problems should support the cycle through which human beings go in solving these problems. Our extensive experience with medical diagnosis, military situation assessment, electronic troubleshooting and other applications, suggest the following cycle which is illustrated in Figure 6. Experimental evidence supporting this description may be found in Eddy and Clanton (1982) Elstein et al (1978) and Zakai et al (1983). Military situation assessment will be used for illustrative purposes. Each step in the cycle represents one type of a decision problem, each of which may require different skills on the part of a human being, and different algorithms on the part of an expert system.

(0) Initial Findings Accumulation
The cycle starts with the presentation of an initial set of specific facts about the situation. These facts may have been observed in the field or they may have been passed to the decision maker, e.g. a G2 officer, through the command channels. They may have come from higher echelons, from parallel units or from subordinate units. They also include indications or responses to information requests that have been placed previously by the G2 and collected by the various information collecting agencies at his disposal.

From thereon the process may be decomposed into the following steps:
(1) Evidence Propagation and Hypothesis Generation
The findings recently obtained are integrated into the existing evidence (which in the first iteration is the apriori information only) and trigger a moving chain of deductions pointing at several alternative interpretations in several perspectives of the battlefield. The uncertainty regarding the truth of these interpretations is updated, and as a result some alternatives may be verified beyond some threshold of condidence, others may be refuted (below some reasonable threshold of confidence) and still others may remain uncertain, though still feasible. At this point an attempt is made to see if the entire puzzle is clear, i.e. if the existing evidence explains the situation in each perspective of the battlefield, and a global interpretation of the situation may be drawn. Those aspects of the battlefield which remain unclear serve as the basis for deriving hypotheses to be worked up in the subsequent stages. The generation of a rich set of plausible hypotheses is the hallmark of a good situation assessor.

(2) Goal(s) Setting for Attention Management
Occasionally — particularly in early stages — too many hypotheses may be triggered by the existing evidence, and not all of them may be simultaneously explored. In such a case goal(s) need to be set on which attention will be focused in the next

25

immediate stages. These may include, for instance, verification/elimination of a specific hypothesis, or differentiation between a group of competing hypotheses. Factors which affect goal determination include the severity and urgency of the candidate alternatives (i.e. enemy attack is expected within 24 hours), their present level of uncertainty and their initial apriori incidence.

(3) Information Sources Evaluation and Selection

Once a goal(s) is set, the information sources which may offer the findings by which this goal may be achieved need to be identified and evaluated. Such an evaluation is based, on one hand, on the potency (information content and reliability) of these information sources to achieve the determined goal, and, on the other hand, on the cost of utilizing them. This cost reflects not only financial, technical and logistic investments, but also the risk involved in getting the information. The information source(s) with the greatest expected contribution to the specified goal(s) compared to its cost is then invoked , e.g. a reconnaissance aircraft. Frequently, a battery of information sources may be utilized simultaneously to permit deeper exploration of a given hypothesis or concurrent exploration of several hypotheses.

(4) Sorting of Evidence by Goals and Hypotheses
As new findings come in, either as a result of the decision maker's request or "voluntarily", they should be sorted with regard to the entire battlefield structure including of course the triggered hypotheses and, on the highest priority, with regard to the current goals. Nevertheless, findings should not be ignored just because they do not contribute directly to the current goal(s) or to the previously activated hypotheses. It is such a lateral thinking that may open new ideas leading to the generation of new hypotheses which may eventually turn out to include the correct ones. Goals need to be set in order to direct effectively the information acquisition path. However, once an indicator is observed, its significance should be analyzed with respect to all of its relevant alternatives.

(5) Evidence Integration
Once all of the relevancy links of the new findings are identified, they are integrated with the existing findings and not just added to them. Recognizing dependencies between new and existing findings may prevent artificial compounding of redundant information. It may also suggest synergy, i.e. the evidence suggested by the group of findings is greater than the sum of the individual findings' evidence. At this stage we may also try to restructure the grouping of findings in an attempt to discover new possible interpretations. The new integrated evidence modifies the uncertainty of existing hypotheses and may suggest new hypotheses concerning the true situation. This completes the cycle and brings us back to stage (1) unless the termination test is positive.

(6) Termination
The situation assessment cycle may be interrupted or fully terminated under one of the following conditions:

(a) A decision may be reached with regard to the true situation in each aspect of the battlefield, all of the (suspicious) findings are explained by this interpretation, and no additional hypotheses are sufficiently triggered to justify further exploration.

(b) Several triggered hypotheses have not yet been settled, however, the cost of removing the remaining uncertainty is relatively high compared to the expected information gain and the impact on the battle plan (or treatment plan if medical diagnosis is the case).

(c) New developments (e.g. sudden enemy attack) force the decision maker to terminate information acquisition and assess the situation as best he can with the existing evidence.

(7) Integrated Summary Composition

The situation assessment process culminates in the composition of the individual decisions made for separate battlefield aspects into one complete and coherent picture that leads to tactical planning. (This is what we called earlier the commitment decisions) The end result is the Intelligence Estimate document which is currently produced manually by the intelligence officer.

5. SUMMARY

We have presented several types of evidential reasoning problems and a detailed description of the Bayesian inference networks (BIN) approach for structuring these problems. The references cited in the paper provide a partial picture of the state of the art in Bayesian evidential reasoning. Much work remains to be done, however, recent developments and experience in this field suggest that the BIN-based approach is a powerful tool for practical expert systems.